\pdfoutput=1

\documentclass[11pt]{article}

\usepackage[svgnames]{xcolor}
\usepackage[final]{acl}

\usepackage{times}
\usepackage{latexsym}

\usepackage[T1]{fontenc}

\usepackage[utf8]{inputenc}

\usepackage{microtype}

\usepackage{inconsolata}

\usepackage{graphicx}

\usepackage{subfig}
\usepackage{pgfplots}
\usepgfplotslibrary{fillbetween}
\usetikzlibrary{patterns}
\usetikzlibrary{pgfplots.groupplots}
\pgfplotsset{compat=1.17}
\usepackage{lscape}
\usepackage{multirow, makecell}
\usepackage{dblfloatfix}
\usepackage{footnote}
\usepackage{graphicx}
\usepackage{tikz}
\usepackage{booktabs, tabularx}
\usepackage{amssymb}
\usepackage{amsmath}
\usepackage{colortbl}

\usepackage{amssymb}
\usepackage{tikz}
\usepackage{cancel}

\title{Evaluating Human Alignment and Model Faithfulness of LLM Rationale}

\author{
    Mohsen Fayyaz$^{1}$ ~ Fan Yin$^{1}$ ~
    \textbf{Jiao Sun$^{2}$}  ~ \textbf{Nanyun Peng$^{1}$} \\
    $^1$ University of California, Los Angeles ~ 
    $^2$ Google \\
    \texttt{\{mohsenfayyaz, fanyin20, violetpeng\}@cs.ucla.edu} ~
    \texttt{jiaosun@google.com}
}

\begin{document}
\maketitle
\begin{abstract}
We study how well large language models (LLMs) explain their generations through rationales -- a set of tokens extracted from the input text that reflect the decision-making process of LLMs.
Specifically, we systematically study rationales derived using two approaches: (1) popular prompting-based methods, where prompts are used to guide LLMs in generating rationales, and (2) technical attribution-based methods, which leverage attention or gradients to identify important tokens.
Our analysis spans three classification datasets with annotated rationales, encompassing tasks with varying performance levels. 
While prompting-based self-explanations are widely used, our study reveals that these explanations are not always as ``aligned'' with the human rationale as attribution-based explanations. Even more so, fine-tuning LLMs to enhance classification task accuracy does not enhance the alignment of prompting-based rationales. Still, it does considerably improve the alignment of attribution-based methods (e.g., Input×Gradient).
More importantly, we show that prompting-based self-explanation is also less ``faithful'' than attribution-based explanations, failing to provide a reliable account of the model's decision-making process. 
To evaluate faithfulness, unlike prior studies that excluded misclassified examples, we evaluate all instances and also examine the impact of fine-tuning and accuracy on alignment and faithfulness. Our findings suggest that inconclusive faithfulness results reported in earlier studies may stem from low classification accuracy.
These findings underscore the importance of more rigorous and comprehensive evaluations of LLM rationales.\footnote{Code and data will be released upon paper acceptance.}

\end{abstract}

\section{Introduction}
The rise of large language models (LLMs) has significantly transformed the field of natural language processing (NLP)~\citep{touvron2023llama, team2023gemini, openai2024gpt4}, enabling a wide range of applications from web question answering to complex reasoning tasks. However, if they cannot clearly and reliably explain their outputs~\citep{ji2023survey}, it limits their deployment in high-stakes scenarios.

\begin{figure}[t]
\centering
    \includegraphics[width=0.45\textwidth, trim=1 25 380 2, clip]{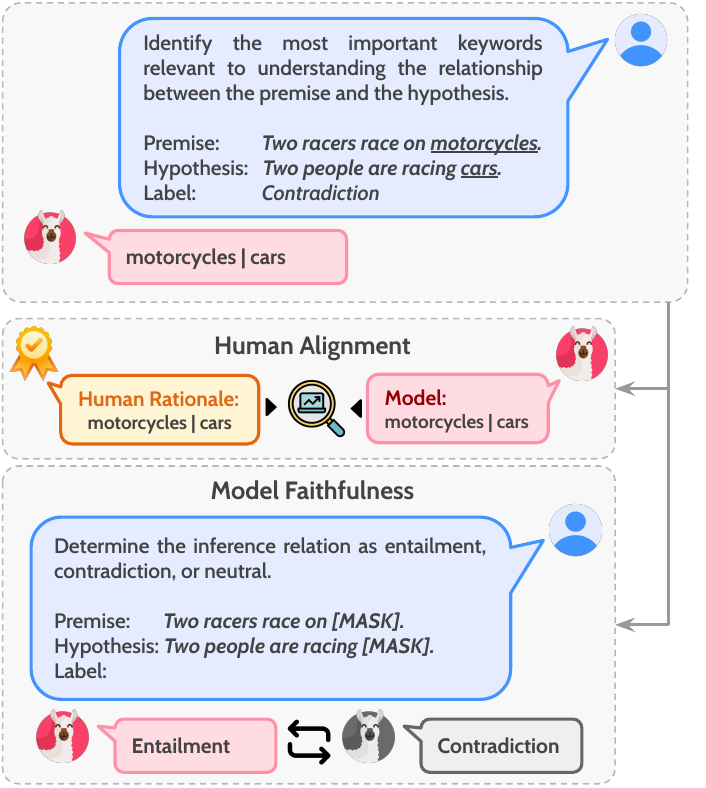}
    \caption{
    An example of our analysis methodology on the e-SNLI dataset. \emph{Human alignment} compares model rationales with human-annotated rationale; \emph{Model faithfulness} measures when model prediction changes (e.g. from \texttt{Contradiction} to \texttt{Entailment}) after masking the rationales identified by the model.\protect\footnotemark
    }
    \label{fig:first_page_diagram}
\end{figure}
\footnotetext{In the first prompt, we use the true label for human alignment, and the predicted label for faithfulness experiments.}


\textit{Rationales}\footnote{Also called \textit{self-explanation} or \textit{extractive rationales} in previous work \citep{huang2023can, madsen2024selfexplanations}.}, i.e., tokens of the input text that are most influential to the models' predictions, are widely studied in the NLP community prior to the era of LLMs to interpret model predictions~\citep{lei-etal-2016-rationalizing, deyoung2019eraser, wiegreffe2019attention, jacovi2020towards}. 
For LLMs, besides attribution-based methods like attention weights~\citep{wiegreffe2019attention} or gradients~\citep{li2016visualizing}, rationales can also be extracted by leveraging the instruction-following ability of LLMs and guiding them with explicit prompts to explain their predictions (Figure~\ref{fig:first_page_diagram}). We call these prompting-based rationales.

To evaluate different rationales, previous works on model interpretation establish two properties of rationales that are critical for successful interpretability: human alignment~\citep{deyoung2019eraser, hase-bansal-2022-models} and faithfulness~\citep{jacovi2020towards}. \textit{Human alignment} refers to
the degree to which the rationales match or align with human-annotated rationales, 
while \textit{faithfulness} assesses whether the rationales truly reflect the model's internal process. 
However, studies on LLM rationales either focus on the faithfulness of off-the-shelf LLMs~\citep{huang2023can, madsen2024selfexplanations}, or their human alignment \cite{chen2023models}, but lack a comprehensive exploration of the two properties together. 
Moreover, they consider LLMs only as out-of-the-box models, without fine-tuning for specific tasks. The impact of fine-tuning to improve task accuracy on the alignment and faithfulness of LLM rationales remains under-explored.

In this paper, we conduct extensive experiments to comprehensively evaluate LLM rationales and bridge the gap in existing research. We consider five state-of-the-art LLMs, encompassing both open-source models (Llama2~\citep{touvron2023llama}, Llama3, Mistral~\citep{jiang2023mistral}) and proprietary models (GPT-3.5-Turbo, GPT-4-Turbo~\citep{openai2024gpt4}). Our study leverages three annotated natural language classification datasets, e-SNLI~\citep{Camburu2018eSNLINL}, FEVER~\citep{thorne-etal-2018-fever} and MedicalBios~\citep{eberle-etal-2023-rather}, to evaluate and compare rationale extraction methods based on prompting strategies and feature attribution-based techniques such as Input×Gradient~\citep{li2016visualizing}.

Through our experiments, we find that attribution-based methods align more closely with humans in most cases, especially after fine-tuning. These methods also show more consistent improvements from fine-tuning compared to prompting.
Additionally, we observe that low classification performance and collapsing predictions are linked to the limitations in evaluating the faithfulness of LLM rationales, shedding light on the task- and model-dependent findings of previous research. Most importantly we demonstrate that attribution-based methods offer a more faithful reflection of the model’s inner workings than prompting.

In summary, our work provides a systematic framework, empirical evidence, and practical recommendations for extracting and evaluating LLM rationales. 
Our findings highlight critical limitations in prompting-based rationales, emphasizing the need for further efforts to improve the interpretability and trustworthiness of LLMs.

\section{Related Work}
\begin{table*}[h]
\centering
\footnotesize
\tabcolsep=0.10cm
\begin{tabular}{lp{5.9cm}p{2.1cm}p{2.1cm}@{\ \ \ \ }p{2.1cm}p{2.1cm}}
\toprule 
  & \textbf{Input} & \textbf{Human} & \textbf{Mistral-7B} & \textbf{LLama-3-8B} & \textbf{GPT-4 Turbo}\\
\midrule 
\multirow{4}{*}{\rotatebox{90}{e-SNLI}}
& Premise: \textbf{Five} children playing soccer chase after a ball. 
\newline Hypothesis: There are \textbf{ten} children playing.
\newline Label: Contradiction
& \textbf{[Five, ten]}
& [\textbf{Five} children, \newline \textbf{ten} children]
& [children, \newline playing]
& [children, \textbf{ten}]

\\\midrule[0.03em]
\multirow{8}{*}{\rotatebox{90}{FEVER}}
& Claim: The Lincoln-Douglas debates happened in Alton, Illinois.
\newline Evidence: The Lincoln -- Douglas Debates -LRB- also known as The Great Debates of 1858 -RRB- were a series of seven debates between Abraham Lincoln ... The debates were held in seven towns in the state of Illinois : Ottawa on August 21 ... \textbf{Alton, on, October, 15} ...
\newline Label: Supports
& \textbf{[Alton, on, \newline October, 15]}
& [Lincoln, \newline Douglas, \newline Illinois, \newline debates]
& [\textbf{Alton}, \newline \textbf{October}, \newline Illinois, \newline Debates]
& [\textbf{Alton}, \newline debates, \newline \textbf{October}, \newline \textbf{15}]

\\\midrule[0.03em]
\multirow{5}{*}{\rotatebox{90}{MedicalBios}}
& His basic \textbf{surgical training} was undertaken in Newcastle. After gaining his \textbf{MRCS} and spending some time carrying out locum SpR work at Hurstwood Park, the Royal London Hospital
\newline Label: Surgeon
& \textbf{[surgical, \newline training, \newline MRCS]}
& [His, \newline \textbf{MRCS}, \newline \textbf{surgical}]
& [\textbf{surgical}, \newline \textbf{training}, \textbf{MRCS}]
& [\textbf{surgical}, \textbf{MRCS}, \newline SpR]



 \\
\bottomrule
\end{tabular}
\caption{
Examples from the datasets are presented, showcasing rationales from human annotations alongside those generated by three models. Human rationales and the corresponding generated words are highlighted in \textbf{bold font}.
} 
\label{tab:examples}
\end{table*}
\label{sec:related}

\paragraph{Interpretability} Recent literature in natural language processing (NLP) has seen a surge in interpretability methods aimed at making models more transparent and understandable. The traditional interpretability methods include 1) attention-based methods, which leverage the attention weights in models like transformers to identify which parts of the input the model focuses on when making a decision~\cite{vaswani2023attention, clark-etal-2019-bert, abnar-zuidema-2020-quantifying}, 
2) Gradient-based methods, which provide explanations by identifying which input tokens most influence the model's output, often using techniques like gradient-based saliency maps~\cite{simonyan2014deep}, or its extension by incorporating the input vector norms or integration~\cite{sundararajan2017axiomatic}. 
3) Vector-based methods that propagate the decomposed representations throughout the model achieving the best faithfulness results on encoder-based models \cite{kobayashi-etal-2020-attention-norm, kobayashi-etal-2021-incorporating-residual,  ferrando-etal-2022-measuring,
modarressi-etal-2022-globenc, modarressi-etal-2023-decompx}.

\paragraph{Rationales} 
Rationales can be categorized as free-form or extractive. Free-form rationales use natural language to explain the model's reasoning, filling in commonsense knowledge gaps. They can improve model performance~\cite{sun-etal-2022-expunations} and user interpretability~\cite{sun-etal-2022-investigating}. Extractive rationales highlight specific parts of the input text that provide sufficient evidence for a prediction, independent of the rest of the input~\cite{lei-etal-2016-rationalizing,deyoung-etal-2020-eraser}. They can also enhance model performance~\cite{huang-etal-2021-exploring,Carton2021WhatTL} and improve human interpretability~\cite{strout-etal-2019-human}. 
Our work focuses on extractive rationales for interpretability evaluation. 
In this research area, \citet{huang2023large} studied faithfulness in ChatGPT, comparing prompting and Lime \cite{ribeiromodel}. \citet{madsen2024selfexplanations} investigated LLM faithfulness on models like Llama2 \cite{touvron2023llama}, Falcon \cite{penedo2023refinedweb}, and Mistral \cite{jiang2023mistral}, noting its dependence on both model and dataset. Despite this, there is still a scarcity of analyses comparing LLM-generated rationales to interpretability methods. To the best of our knowledge, no studies seem to have assessed human alignment and model faithfulness jointly or used fine-tuning to overcome faithfulness evaluation limitations and explore its effects. 

\section{Experimental Setup}

\subsection{Datasets}
We use three natural language classification datasets, each annotated with human rationales that highlight the key input words essential for determining the correct label. Table~\ref{tab:examples} shows examples of these datasets alongside the human rationale annotation and model-generated rationale.

\noindent{\bf e-SNLI}
This dataset \cite{camburu2018snli, deyoung2019eraser} is a natural language inference task with three classes including Entailment, Contradiction, and Neutral, showing the relation between the premise and hypothesis sentences.
We utilize 5,000 examples from the training set and 300 examples from the test set.

\noindent{\bf FEVER}
The Fact Extraction and Verification dataset \cite{thorne2018fever, deyoung2019eraser} focuses on verifying claims based on provided evidence. Each claim is labeled as either "supports" or "refutes," with an annotated rationale indicating the specific portion of the evidence that underpins this classification. We use 5,000 samples from the training set and 300 from the test set.

\noindent{\bf MedicalBios}
\cite{eberle-etal-2023-rather} consists of human rationale annotations for a subset of 100 samples (five medical classes) from the BIOS dataset \cite{BIOS} for the occupation classification task.

\begin{figure*}[t!]
\centering

    \centering
    \includegraphics[width=.32\textwidth]{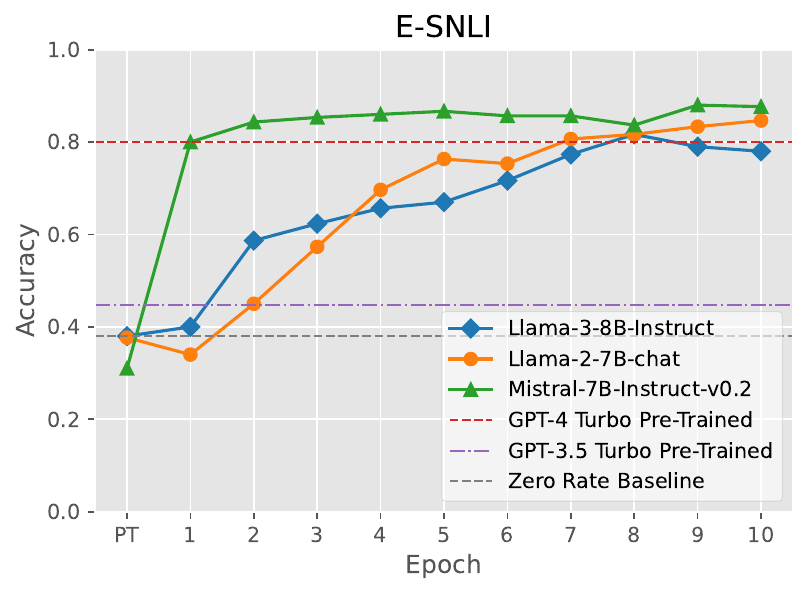}\hfill
    \includegraphics[width=.32\textwidth]{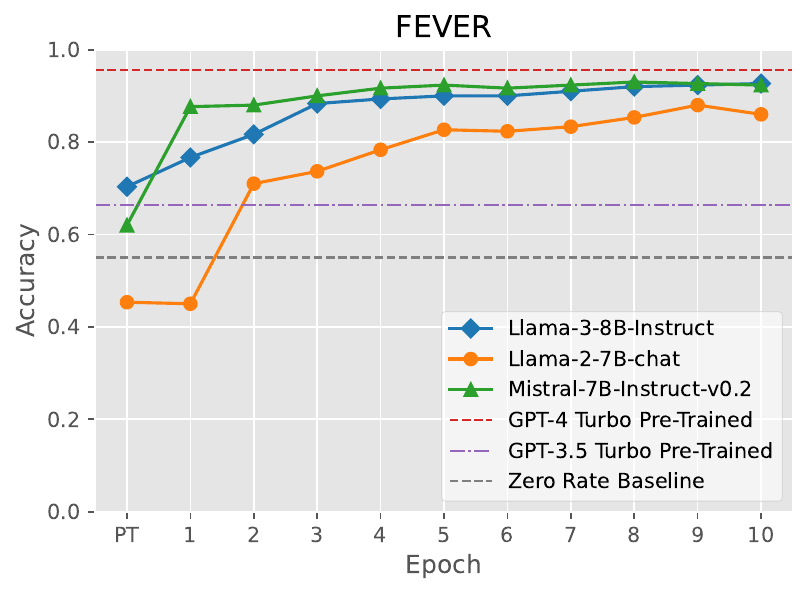}\hfill
    \includegraphics[width=.32\textwidth]{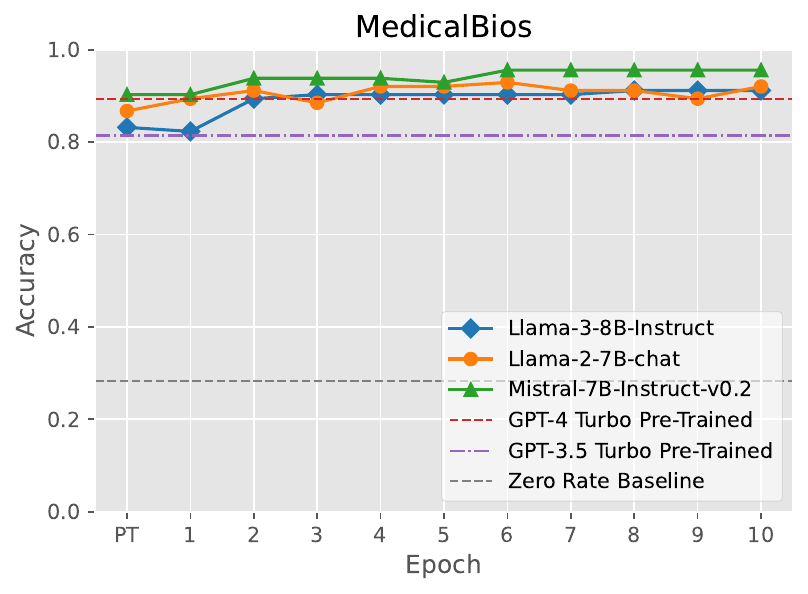}
    \caption{
    Classification accuracy throughout 10 epochs of fine-tuning. PT denotes the pre-trained model's accuracy. "Zero Rate Baseline" refers to the performance of a classifier that assigns all inputs to the majority class. Our tasks include E-SNLI and FEVER, where pre-trained models tend to underperform, as well as MedicalBios, where they demonstrate strong performance.
    }
    \label{fig:accuracy_epochs_horizontal}
\end{figure*}

\subsection{Models}
We employ five of the latest LLMs. From the open-source models, we utilize Llama2 \cite{touvron2023llama}, LLama3, and Mistral \cite{jiang2023mistral}. For proprietary models, we include GPT3.5-Turbo and GPT4-Turbo \cite{openai2024gpt4}. All models are prompted without sampling during generation, leading to deterministic outputs (See Table~\ref{tab:models} for model information).

\subsection{Methods}
\subsubsection{Prompting-Based Method}
We employ different prompts for each stage of our experiments which are shown in Tables~\ref{tab:prompts_esnli}, \ref{tab:prompts_fever}, and \ref{tab:prompts_medicalbios}. Our prompts are engineered to convey the necessary information in a few sentences without confusing the models.
We also experiment with two variations of each explanation prompt to manage the number of words the model generates.

\noindent{\bf Unbound Prompt}
In this method, the model autonomously determines the appropriate length of its generated text without word count restrictions.

\noindent{\bf Top-Var Prompt}
In this prompt, the model generates the top-$k$ words, where $k$ matches the exact number of words annotated in the human rationales for each sentence. This method controls for word count in our experiments, allowing us to evaluate model alignment independently of its word importance threshold and ensuring a fair evaluation of faithfulness by keeping the number of masked words consistent across experiments.

\subsubsection{Attribution-Based Methods}
We employ the Inseq library \cite{sarti-etal-2023-inseq} to implement attribution-based methods for LLMs. Specifically, we select three available options: (i) Attention Weight Attribution, which utilizes the model's internal attention weights \cite{wiegreffe2019attention}; (ii) Simple Gradients (Saliency), which is based on the gradients of the output with respect to the inputs \cite{Simonyan2014DeepIC}; and (iii) Input×Gradient, which factors in both the input vector size and the gradient in its calculations \cite{li2016visualizing}. We choose these methods because of their demonstrated faithfulness in previous work on NLP models \cite{atanasova-etal-2020-diagnostic, modarressi-etal-2022-globenc, modarressi-etal-2023-decompx}, and their potential for efficient execution on large language models with limited computational resources. For each method, we focus on the output label word produced by the model in response to the classification prompt (Table~\ref{tab:prompts_esnli}), calculating its attributions to the input tokens.

\newcommand{\grayrule}{\arrayrulecolor{black!30}\midrule[0.01pt]\arrayrulecolor{black}}
\newcommand{\grayruletwoseven}{\arrayrulecolor{black!30}\cmidrule{2-6}\arrayrulecolor{black}}
\newcommand{\grayruletwonine}{\arrayrulecolor{black!30}\cmidrule{2-9}\arrayrulecolor{black}}
\definecolor{attrcolor}{RGB}{237, 235, 240}
\definecolor{promptcolor}{RGB}{252, 242, 230}

\begin{table*}[t!]
\begin{center}
\small
\tabcolsep=0.09cm

\begin{tabular}{cll|ccc|ccc}
\toprule
 &  &  & \multicolumn{3}{c|}{\textbf{Pre-trained Model F1}} & \multicolumn{3}{c}{\textbf{Fine-tuned Model F1}} \\
 &  &  & E-SNLI & FEVER & MedBios & E-SNLI & FEVER & MedicalBios \\
Model & Method & Selection &  &  &  &  &  &  \\
\midrule
\multirow[c]{5}{*}{\shortstack{\textbf{Mistral-7B} \\ \textbf{Instruct-v0.2}}} & \multirow[c]{2}{*}{\textsc{Prompting}} & \textsc{Unbound} & \cellcolor{promptcolor!75} 36.36 & \cellcolor{promptcolor!75} 24.68 & \cellcolor{promptcolor!75} 38.14 & \cellcolor{promptcolor!75} 33.03 (\textbf{\textcolor{DarkRed}{\textminus 3.33}}) & \cellcolor{promptcolor!75} 25.64 (\textcolor{blue}{+0.95}) & \cellcolor{promptcolor!75} 38.87 (\textcolor{blue}{+0.72}) \\
\textbf{} & \textbf{} & \textsc{Top-Var} & \cellcolor{promptcolor!75} 40.08 & \cellcolor{promptcolor!75} 26.24 & \cellcolor{promptcolor!75} \underline{44.11} & \cellcolor{promptcolor!75} 35.29 (\textbf{\textcolor{DarkRed}{\textminus 4.79}}) & \cellcolor{promptcolor!75} 26.45 (\textcolor{blue}{+0.21}) & \cellcolor{promptcolor!75} 45.02 (\textcolor{blue}{+0.92}) \\

\grayruletwonine \textbf{} & \textsc{Attention} & \textsc{Top-Var} & \cellcolor{attrcolor!75} 36.26 & \cellcolor{attrcolor!75} 37.23 & \cellcolor{attrcolor!75} 37.36 & \cellcolor{attrcolor!75} 42.14 (\textbf{\textcolor{blue}{+5.87}}) & \cellcolor{attrcolor!75} 38.64 (\textcolor{blue}{+1.41}) & \cellcolor{attrcolor!75} 38.29 (\textcolor{blue}{+0.93}) \\

\textbf{} & \textsc{Saliency} & \textsc{Top-Var} & \cellcolor{attrcolor!75} \underline{46.46} & \cellcolor{attrcolor!75} \underline{37.96} & \cellcolor{attrcolor!75} 40.76 & \cellcolor{attrcolor!75} \underline{49.69} (\textbf{\textcolor{blue}{+3.24}}) & \cellcolor{attrcolor!75} \underline{40.44} (\textbf{\textcolor{blue}{+2.47}}) & \cellcolor{attrcolor!75} \underline{46.91} (\textbf{\textcolor{blue}{+6.15}}) \\

\textbf{} & \textsc{Input×Grad} & \textsc{Top-Var} & \cellcolor{attrcolor!75} 40.45 & \cellcolor{attrcolor!75} 36.36 & \cellcolor{attrcolor!75} 40.10 & \cellcolor{attrcolor!75} 42.70 (\textbf{\textcolor{blue}{+2.25}}) & \cellcolor{attrcolor!75} 39.21 (\textbf{\textcolor{blue}{+2.85}}) & \cellcolor{attrcolor!75} 45.38 (\textbf{\textcolor{blue}{+5.28}}) \\
\midrule 
\multirow[c]{5}{*}{\shortstack{\textbf{Llama-2-7b} \\ \textbf{chat}}} & \multirow[c]{2}{*}{\textsc{Prompting}} & \textsc{Unbound} & \cellcolor{promptcolor!75} 38.57 & \cellcolor{promptcolor!75} 14.69 & \cellcolor{promptcolor!75} 32.64 & \cellcolor{promptcolor!75} 38.44 (\textcolor{DarkRed}{\textminus 0.13}) & \cellcolor{promptcolor!75} 14.19 (\textcolor{DarkRed}{\textminus 0.50}) & \cellcolor{promptcolor!75} 33.87 (\textcolor{blue}{+1.22}) \\
\textbf{} & \textbf{} & \textsc{Top-Var} & \cellcolor{promptcolor!75} \underline{45.65} & \cellcolor{promptcolor!75} 20.89 & \cellcolor{promptcolor!75} \underline{49.77} & \cellcolor{promptcolor!75} 44.33 (\textcolor{DarkRed}{\textminus 1.31}) & \cellcolor{promptcolor!75} 20.43 (\textcolor{DarkRed}{\textminus 0.46}) & \cellcolor{promptcolor!75} \underline{50.48} (\textcolor{blue}{+0.71}) \\

\grayruletwonine \textbf{} & \textsc{Attention} & \textsc{Top-Var} & \cellcolor{attrcolor!75} 31.65 & \cellcolor{attrcolor!75} \underline{32.16} & \cellcolor{attrcolor!75} 31.79 & \cellcolor{attrcolor!75} 32.90 (\textcolor{blue}{+1.25}) & \cellcolor{attrcolor!75} 32.94 (\textcolor{blue}{+0.78}) & \cellcolor{attrcolor!75} 32.33 (\textcolor{blue}{+0.54}) \\

\textbf{} & \textsc{Saliency} & \textsc{Top-Var} & \cellcolor{attrcolor!75} 34.92 & \cellcolor{attrcolor!75} 30.52 & \cellcolor{attrcolor!75} 36.49 & \cellcolor{attrcolor!75} \underline{46.56} (\textbf{\textcolor{blue}{+11.64}}) & \cellcolor{attrcolor!75} \underline{38.52} (\textbf{\textcolor{blue}{+8.00}}) & \cellcolor{attrcolor!75} 34.12 (\textbf{\textcolor{DarkRed}{\textminus 2.37}}) \\

\textbf{} & \textsc{Input×Grad} & \textsc{Top-Var} & \cellcolor{attrcolor!75} 35.43 & \cellcolor{attrcolor!75} 31.01 & \cellcolor{attrcolor!75} 37.84 & \cellcolor{attrcolor!75} 46.38 (\textbf{\textcolor{blue}{+10.96}}) & \cellcolor{attrcolor!75} 38.05 (\textbf{\textcolor{blue}{+7.03}}) & \cellcolor{attrcolor!75} 35.35 (\textbf{\textcolor{DarkRed}{\textminus 2.49}}) \\
\midrule 
\multirow[c]{5}{*}{\shortstack{\textbf{Llama-3-8B} \\ \textbf{Instruct}}} & \multirow[c]{2}{*}{\textsc{Prompting}} & \textsc{Unbound} & \cellcolor{promptcolor!75} 37.68 & \cellcolor{promptcolor!75} 23.79 & \cellcolor{promptcolor!75} 46.52 & \cellcolor{promptcolor!75} 28.14 (\textbf{\textcolor{DarkRed}{\textminus 9.53}}) & \cellcolor{promptcolor!75} 21.83 (\textbf{\textcolor{DarkRed}{\textminus 1.96}}) & \cellcolor{promptcolor!75} 47.94 (\textcolor{blue}{+1.42}) \\
\textbf{} & \textbf{} & \textsc{Top-Var} & \cellcolor{promptcolor!75} 43.07 & \cellcolor{promptcolor!75} 28.76 & \cellcolor{promptcolor!75} \underline{59.99} & \cellcolor{promptcolor!75} 30.03 (\textbf{\textcolor{DarkRed}{\textminus 13.04}}) & \cellcolor{promptcolor!75} 32.58 (\textbf{\textcolor{blue}{+3.82}}) & \cellcolor{promptcolor!75} \underline{59.76} (\textcolor{DarkRed}{\textminus 0.23}) \\

\grayruletwonine \textbf{} & \textsc{Attention} & \textsc{Top-Var} & \cellcolor{attrcolor!75} 28.67 & \cellcolor{attrcolor!75} 33.87 & \cellcolor{attrcolor!75} 26.48 & \cellcolor{attrcolor!75} 28.60 (\textcolor{DarkRed}{\textminus 0.07}) & \cellcolor{attrcolor!75} 34.80 (\textcolor{blue}{+0.93}) & \cellcolor{attrcolor!75} 29.57 (\textbf{\textcolor{blue}{+3.10}}) \\

\textbf{} & \textsc{Saliency} & \textsc{Top-Var} & \cellcolor{attrcolor!75} 41.42 & \cellcolor{attrcolor!75} 34.29 & \cellcolor{attrcolor!75} 39.68 & \cellcolor{attrcolor!75} 44.62 (\textbf{\textcolor{blue}{+3.20}}) & \cellcolor{attrcolor!75} 35.61 (\textcolor{blue}{+1.32}) & \cellcolor{attrcolor!75} 41.67 (\textbf{\textcolor{blue}{+1.98}}) \\

\textbf{} & \textsc{Input×Grad} & \textsc{Top-Var} & \cellcolor{attrcolor!75} \underline{44.59} & \cellcolor{attrcolor!75} \underline{35.27} & \cellcolor{attrcolor!75} 45.61 & \cellcolor{attrcolor!75} \underline{49.22} (\textbf{\textcolor{blue}{+4.63}}) & \cellcolor{attrcolor!75} \underline{36.30} (\textcolor{blue}{+1.04}) & \cellcolor{attrcolor!75} 46.68 (\textcolor{blue}{+1.07}) \\
\midrule 
\multirow[c]{2}{*}{\shortstack{\textbf{GPT-3.5 Turbo} \\ \textbf{1106}}} & \multirow[c]{2}{*}{\textsc{Prompting}} & \textsc{Unbound} & \cellcolor{promptcolor!75} 43.14 & \cellcolor{promptcolor!75} 22.76 & \cellcolor{promptcolor!75} 42.95 & \cellcolor{promptcolor!75} - & \cellcolor{promptcolor!75} - & \cellcolor{promptcolor!75} - \\
\textbf{} & \textbf{} & \textsc{Top-Var} & \cellcolor{promptcolor!75} 46.06 & \cellcolor{promptcolor!75} 34.27 & \cellcolor{promptcolor!75} 53.96 & \cellcolor{promptcolor!75} - & \cellcolor{promptcolor!75} - & \cellcolor{promptcolor!75} - \\
\midrule 
\multirow[c]{2}{*}{\shortstack{\textbf{GPT-4 Turbo} \\ \textbf{2024-04-09}}} & \multirow[c]{2}{*}{\textsc{Prompting}} & \textsc{Unbound} & \cellcolor{promptcolor!75} 51.44 & \cellcolor{promptcolor!75} 29.48 & \cellcolor{promptcolor!75} 53.25 & \cellcolor{promptcolor!75} - & \cellcolor{promptcolor!75} - & \cellcolor{promptcolor!75} - \\
\textbf{} & \textbf{} & \textsc{Top-Var} & \cellcolor{promptcolor!75} \textbf{52.53} & \cellcolor{promptcolor!75} \textbf{43.23} & \cellcolor{promptcolor!75} \textbf{60.77} & \cellcolor{promptcolor!75} - & \cellcolor{promptcolor!75} - & \cellcolor{promptcolor!75} - \\
\midrule 
\multirow[c]{1}{*}{\textbf{Random}} & \multirow[c]{1}{*}{\textsc{Random}} & \textsc{Top-Var} &  26.54 &  33.90 & 21.97 & 26.54 &  33.90 & 21.97 \\
\bottomrule
\end{tabular}

\end{center}
\caption{
Human alignment \textsc{F1$\uparrow$} score. The difference in alignment between the fine-tuned (10 epochs) and pre-trained model is reported in the parentheses. Average random baseline (Selecting Top-Var random words) over 100 seeds is also reported. The highest alignment in each combination of model and dataset is \underline{underlined}. Attribution-based methods are more aligned with humans in the majority of the cases, especially after fine-tuning. They also exhibit more consistent improvements by fine-tuning compared to prompting.
}
\label{tab:human_alignment}
\end{table*}

\section{Results}
In this section, we delve into utilizing both prompting-based and attribution-based approaches to extract rationale from the model, focusing on two aspects: human alignment and model faithfulness. Furthermore, we conduct fine-tuning experiments on open LLMs to examine how task performance influences alignment and faithfulness.

\subsection{Task Performance}

We begin our results by examining the accuracy of the models on the datasets. Figure~\ref{fig:accuracy_epochs_horizontal} presents the pre-trained (PT) off-the-shelf accuracy of the models, along with their performance improvements during fine-tuning for 10 epochs.

As described by \citet{wang-etal-2024-rethinking} and \citet{zhong2023chatgpt}, LLMs may underperform small fine-tuned models such as BERT, and smaller LLMs like LLaMA-2-7B might even collapse entirely. \citet{madsen2024selfexplanations} reports similar behavior, with certain combinations of tasks and models performing poorly across different prompt variations. This pattern is evident in our results as well, where pre-trained open LLMs exhibit near-random accuracy on the e-SNLI and FEVER datasets (Figure~\ref{fig:accuracy_epochs_horizontal}).

Unlike previous work \cite{madsen2024selfexplanations} that simply excluded misclassified examples from their faithfulness evaluations—which may represent more than half of the dataset—we include all examples in our analysis. Additionally, we fine-tune\footnote{The hyperparameters are provided in Table~\ref{tab:finetuning_hyperparameters}.} the LLMs using LoRA \cite{hu2022lora}. As shown in Figure~\ref{fig:accuracy_epochs_horizontal}, the classification performance of LLaMA-2, LLaMA-3, and Mistral improves significantly after fine-tuning on the underperforming datasets, e-SNLI and FEVER, while also showing slight enhancements on the well-performing MedicalBios dataset. This fine-tuning narrows the performance gap with GPT-4-Turbo across all cases.

In the following sections, we will explore the alignment and faithfulness of pre-trained and fine-tuned models and investigate how task performance influences these aspects.

\subsection{Human Alignment}

The human-annotated rationales provide explanations for the ground truth label. Therefore, we first ask the model to generate its rationale for the true label (by prompting or attribution). Then to measure alignment, we compute the F1 score, which balances precision (the proportion of correctly generated words out of all generated words) and recall (the proportion of correctly generated words out of all relevant words). This score compares the model's generated rationales with human-annotated ones, assessing how well the model aligns with human explanations. Table~\ref{tab:human_alignment} presents the F1 scores for the pre-trained and fine-tuned (epoch 10) models.

First, comparing the evaluated models reveals GPT-4-Turbo to be the most aligned with humans. Although the other models show varying levels of alignment depending on the task, GPT-3.5-Turbo and Llama3 often provide better prompting explanations than Llama2 and Mistral.

Second, we note that providing additional information about the number of words selected by humans in \textit{Top-Var} settings enhances alignment, indicating disparities between model thresholds for word importance in \textit{Unbound} prompting compared to human annotators.

Third, attribution-based methods tend to align more closely with human reasoning than prompting-based methods, especially after fine-tuning.
This suggests that the model may be attending to words in a manner similar to human thought processes, although this cannot be fully explained through language modeling. Furthermore, among the attribution-based methods, Input×Gradient and Saliency appear to outperform raw attention weights, which is anticipated based on existing literature \cite{ferrando-etal-2022-measuring}.

\begin{figure*}[t!]
\centering

    \centering
    \includegraphics[width=.32\textwidth]{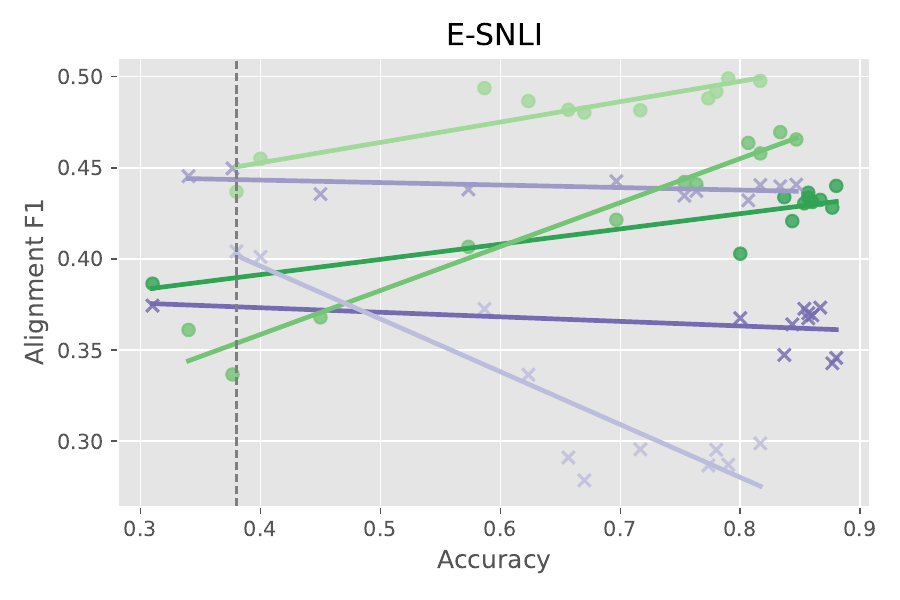}\hfill
    \includegraphics[width=.32\textwidth]{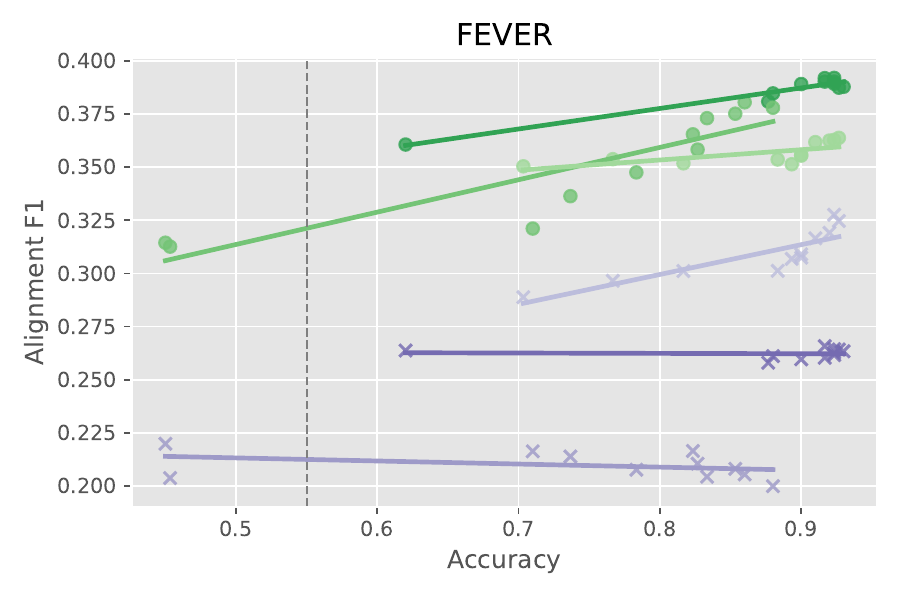}\hfill
    \includegraphics[width=.32\textwidth]{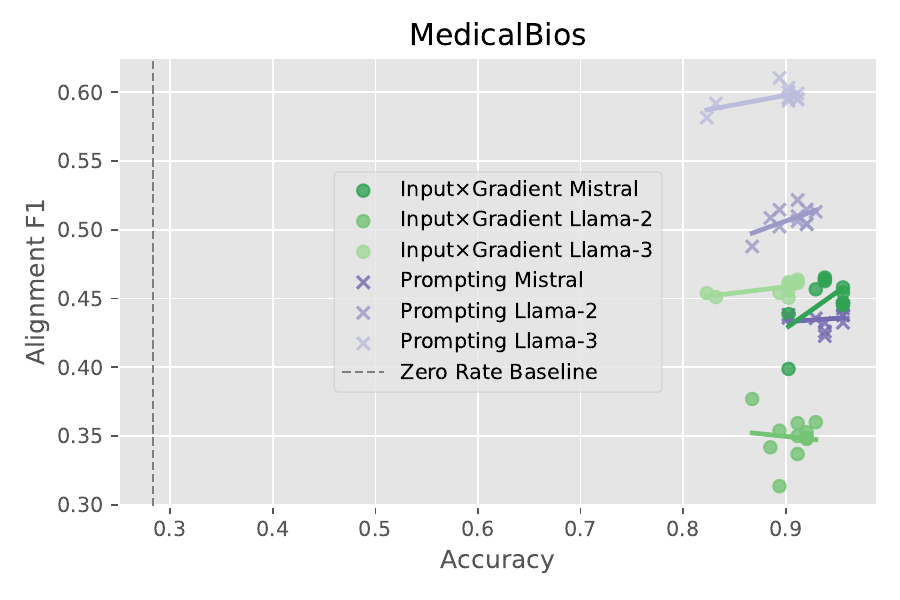}
    \captionsetup{aboveskip=4pt}
    \caption{
    Accuracy and Human Alignment F1 score of 10 epochs of fine-tuning.
    }
    \label{fig:acc_alignment_corr}
\end{figure*}
\newcolumntype{\pc}{>{\columncolor{promptcolor}}r}
\newcolumntype{\ac}{>{\columncolor{attrcolor}}r}

\begin{table}[t!]
\begin{center}
\small

\tabcolsep=0.11cm

\begin{tabular}{cl\ac\pc}
\toprule
 & Method & Input×Gradient & Prompting \\
Dataset & Model &  &  \\
\midrule
\midrule \multirow[c]{3}{*}{\text{E-SNLI}} & 
\text{Mistral-7B} & 0.85 (0.00) & -0.35 (0.30) \\
\text{} & \text{Llama-2-7b} & 0.98 (0.00) & -0.52 (0.10) \\
\text{} & \text{Llama-3-8B} & 0.88 (0.00) & -0.90 (0.00) \\

\midrule \multirow[c]{3}{*}{\text{FEVER}} & 
\text{Mistral-7B} & 0.97 (0.00) & -0.06 (0.86) \\
\text{} & \text{Llama-2-7b} & 0.90 (0.00) & -0.36 (0.28) \\
\text{} & \text{Llama-3-8B} & 0.70 (0.02) & 0.86 (0.00) \\

\midrule \multirow[c]{3}{*}{\text{MedicalBios}} 
& \text{Mistral-7B} & 0.58 (0.06) & 0.16 (0.64) \\ 
\text{} & \text{Llama-2-7b} & -0.09 (0.79) & 0.57 (0.07) \\
\text{} & \text{Llama-3-8B} & 0.61 (0.04) & 0.61 (0.05) \\

\bottomrule
\end{tabular}

\end{center}
\captionsetup{aboveskip=3pt}
\caption{
The Pearson Correlation (p-value) of Human alignment \textsc{F1$\uparrow$} score and classification accuracy across 10 epochs of fine-tuning each model. Input×Gradient is highly correlated with accuracy while prompting doesn't show a clear trend.
}
\vspace{-0.2in}
\label{tab:acc_alignment_corr}
\end{table}

\subsection{Effect of Classification Performance on Alignment}

To analyze the effect of task performance on alignment more comprehensively, we rerun the alignment experiments for ``Prompt Top-Var'' and ``Input×Gradient Top-Var'' across all 10 epochs, as shown in Figure~\ref{fig:acc_alignment_corr}. 
Moreover, Table~\ref{tab:acc_alignment_corr} shows the correlation between the Alignment F1 and the model's classification accuracy.

The results suggest a general trend of improved alignment for attribution-based methods with increasing accuracies throughout epochs.\footnote{With the exception of MedicalBios, which does not show a significant range of accuracy improvement (Less than 10\%), making it not a great candidate for this analysis.} This enhancement can stem from their reliance on the classification prompt and the classification capabilities of LLMs, which may fall short in off-the-shelf models. Consequently, fine-tuning enhances performance and affects the model's internal processes, potentially making them more accessible and transparent for detection via attention or gradient-based methods. Moreover, fine-tuning can guide the model's attention to the correct words, especially in datasets like e-SNLI where pre-trained classification accuracy was low.\footnote{Figure~\ref{fig:token_importance_ft} illustrates qualitative examples of such cases.} As a result, these gradient-based methods can identify more human-aligned rationales by tracing back the attributions from the output label to the input sentence in fine-tuned models. Nonetheless, in prompting methods, this improvement in classification seems to function independently of the explanation task, as it does not demonstrate clear or significant enhancements.

\begin{table*}[t!]
\begin{center}
\small
\setlength{\tabcolsep}{0.7mm}

\begin{tabular}{lll|ccc|rrr}
\toprule
 &  &  & \multicolumn{3}{c|}{Pre-trained Faithfulness} & \multicolumn{3}{c}{Fine-tuned Faithfulness} \\
 &  &  & \multicolumn{1}{c}{E-SNLI} & \multicolumn{1}{c}{FEVER} & \multicolumn{1}{c|}{MedBios} & \multicolumn{1}{c}{E-SNLI} & \multicolumn{1}{c}{FEVER} & \multicolumn{1}{c}{MedicalBios} \\
Model & Method &  &  &  &  &  &  &  \\
\midrule
\multirow[c]{7}{*}{\shortstack{\textbf{Mistral-7B} \\ \textbf{Instruct-v0.2}}} & \textsc{Prompting} & \textsc{} & \cellcolor{promptcolor!75} \textbf{\textcolor{DarkRed}{1.00}} & \cellcolor{promptcolor!75} \textbf{\textcolor{DarkRed}{1.70}} & \cellcolor{promptcolor!75} 19.27 & \cellcolor{promptcolor!75} 16.33 (\textbf{\textcolor{blue}{+15.33}}) & \cellcolor{promptcolor!75} 7.33 (\textcolor{blue}{+5.63}) & \cellcolor{promptcolor!75} 8.85 (\textbf{\textcolor{DarkRed}{\textminus 10.42}}) \\

\grayruletwonine \textbf{} & \textsc{Attention} & \textsc{} & \cellcolor{attrcolor!75} \textbf{\textcolor{DarkRed}{0.00}} & \cellcolor{attrcolor!75} 15.65 & \cellcolor{attrcolor!75} 24.77 & \cellcolor{attrcolor!75} 43.67 (\textbf{\textcolor{blue}{+43.67}}) & \cellcolor{attrcolor!75} 11.33 (\textcolor{DarkRed}{\textminus 4.31}) & \cellcolor{attrcolor!75} 12.39 (\textbf{\textcolor{DarkRed}{\textminus 12.38}}) \\

\textbf{} & \textsc{Saliency} & \textsc{} & \cellcolor{attrcolor!75} \textbf{\textcolor{DarkRed}{0.00}} & \cellcolor{attrcolor!75} 14.97 & \cellcolor{attrcolor!75} 30.28 & \cellcolor{attrcolor!75} \underline{50.00} (\textbf{\textcolor{blue}{+50.00}}) & \cellcolor{attrcolor!75} \underline{12.33} (\textcolor{DarkRed}{\textminus 2.63}) & \cellcolor{attrcolor!75} 15.93 (\textbf{\textcolor{DarkRed}{\textminus 14.35}}) \\

\textbf{} & \textsc{Input×Grad} & \textsc{} & \cellcolor{attrcolor!75} \textbf{\textcolor{DarkRed}{0.00}} & \cellcolor{attrcolor!75} 14.29 & \cellcolor{attrcolor!75} 28.44 & \cellcolor{attrcolor!75} 41.33 (\textbf{\textcolor{blue}{+41.33}}) & \cellcolor{attrcolor!75} 11.67 (\textcolor{DarkRed}{\textminus 2.62}) & \cellcolor{attrcolor!75} \underline{16.81} (\textbf{\textcolor{DarkRed}{\textminus 11.63}}) \\

\grayruletwonine \textbf{} & \textsc{Random} & \textsc{} &  \textbf{\textcolor{DarkRed}{0.33}} &  12.24 &  5.50 &  27.00 (\textbf{\textcolor{blue}{+26.67}}) &  5.00 (\textcolor{DarkRed}{\textminus 7.24}) &  2.65 (\textcolor{DarkRed}{\textminus 2.85}) \\

\textbf{} & \textsc{Human} & \textsc{} &  \textbf{\textcolor{DarkRed}{0.67}} &  \textbf{\textcolor{DarkRed}{1.70}} &  44.04 &  50.00 (\textbf{\textcolor{blue}{+49.33}}) &  18.67 (\textbf{\textcolor{blue}{+16.97}}) &  14.16 (\textbf{\textcolor{DarkRed}{\textminus 29.88}}) \\

\textbf{} & \textsc{All} & \textsc{} &  \textbf{\textcolor{DarkRed}{0.00}} &  \textbf{\textcolor{DarkRed}{1.70}} &  100.00 &  68.67 (\textbf{\textcolor{blue}{+68.67}}) &  23.00 (\textbf{\textcolor{blue}{+21.30}}) &  84.96 (\textbf{\textcolor{DarkRed}{\textminus 15.04}}) \\
\midrule 
\multirow[c]{7}{*}{\shortstack{\textbf{Llama-2-7b} \\ \textbf{chat}}} & \textsc{Prompting} & \textsc{} & \cellcolor{promptcolor!75} \textbf{\textcolor{DarkRed}{1.34}} & \cellcolor{promptcolor!75} \textbf{\textcolor{DarkRed}{0.00}} & \cellcolor{promptcolor!75} 20.56 & \cellcolor{promptcolor!75} 33.00 (\textbf{\textcolor{blue}{+31.66}}) & \cellcolor{promptcolor!75} 8.33 (\textcolor{blue}{+8.33}) & \cellcolor{promptcolor!75} 14.16 (\textcolor{DarkRed}{\textminus 6.40}) \\

\grayruletwonine \textbf{} & \textsc{Attention} & \textsc{} & \cellcolor{attrcolor!75} \textbf{\textcolor{DarkRed}{0.67}} & \cellcolor{attrcolor!75} \textbf{\textcolor{DarkRed}{4.00}} & \cellcolor{attrcolor!75} 25.23 & \cellcolor{attrcolor!75} 31.33 (\textbf{\textcolor{blue}{+30.66}}) & \cellcolor{attrcolor!75} 9.00 (\textcolor{blue}{+5.00}) & \cellcolor{attrcolor!75} 14.16 (\textbf{\textcolor{DarkRed}{\textminus 11.07}}) \\

\textbf{} & \textsc{Saliency} & \textsc{} & \cellcolor{attrcolor!75} \textbf{\textcolor{DarkRed}{1.68}} & \cellcolor{attrcolor!75} \textbf{\textcolor{DarkRed}{4.33}} & \cellcolor{attrcolor!75} 37.38 & \cellcolor{attrcolor!75} \underline{46.67} (\textbf{\textcolor{blue}{+44.99}}) & \cellcolor{attrcolor!75} \underline{12.33} (\textcolor{blue}{+8.00}) & \cellcolor{attrcolor!75} \underline{16.81} (\textbf{\textcolor{DarkRed}{\textminus 20.57}}) \\

\textbf{} & \textsc{Input×Grad} & \textsc{} & \cellcolor{attrcolor!75} \textbf{\textcolor{DarkRed}{2.01}} & \cellcolor{attrcolor!75} \textbf{\textcolor{DarkRed}{4.33}} & \cellcolor{attrcolor!75} 37.38 & \cellcolor{attrcolor!75} 46.00 (\textbf{\textcolor{blue}{+43.99}}) & \cellcolor{attrcolor!75} 11.33 (\textcolor{blue}{+7.00}) & \cellcolor{attrcolor!75} \underline{16.81} (\textbf{\textcolor{DarkRed}{\textminus 20.57}}) \\

\grayruletwonine \textbf{} & \textsc{Random} & \textsc{} &  \textbf{\textcolor{DarkRed}{1.34}} &  \textbf{\textcolor{DarkRed}{3.33}} &  15.89 &  26.67 (\textbf{\textcolor{blue}{+25.32}}) &  7.00 (\textcolor{blue}{+3.67}) &  5.31 (\textbf{\textcolor{DarkRed}{\textminus 10.58}}) \\

\textbf{} & \textsc{Human} & \textsc{} &  \textbf{\textcolor{DarkRed}{1.01}} &  \textbf{\textcolor{DarkRed}{0.00}} &  50.47 &  50.00 (\textbf{\textcolor{blue}{+48.99}}) &  14.33 (\textbf{\textcolor{blue}{+14.33}}) &  20.35 (\textbf{\textcolor{DarkRed}{\textminus 30.11}}) \\

\textbf{} & \textsc{All} & \textsc{} &  \textbf{\textcolor{DarkRed}{0.00}} &  \textbf{\textcolor{DarkRed}{0.00}} &  71.03 &  57.33 (\textbf{\textcolor{blue}{+57.33}}) &  19.33 (\textbf{\textcolor{blue}{+19.33}}) &  72.57 (\textcolor{blue}{+1.54}) \\
\midrule 
\multirow[c]{7}{*}{\shortstack{\textbf{Llama-3-8B} \\ \textbf{Instruct}}} & \textsc{Prompting} & \textsc{} & \cellcolor{promptcolor!75} 15.33 & \cellcolor{promptcolor!75} \textbf{\textcolor{DarkRed}{1.33}} & \cellcolor{promptcolor!75} 36.94 & \cellcolor{promptcolor!75} 22.33 (\textcolor{blue}{+7.00}) & \cellcolor{promptcolor!75} 11.33 (\textcolor{blue}{+10.00}) & \cellcolor{promptcolor!75} 16.07 (\textbf{\textcolor{DarkRed}{\textminus 20.87}}) \\

\grayruletwonine \textbf{} & \textsc{Attention} & \textsc{} & \cellcolor{attrcolor!75} 21.33 & \cellcolor{attrcolor!75} 17.33 & \cellcolor{attrcolor!75} 21.62 & \cellcolor{attrcolor!75} 29.67 (\textcolor{blue}{+8.33}) & \cellcolor{attrcolor!75} 11.33 (\textcolor{DarkRed}{\textminus 6.00}) & \cellcolor{attrcolor!75} 10.71 (\textbf{\textcolor{DarkRed}{\textminus 10.91}}) \\

\textbf{} & \textsc{Saliency} & \textsc{} & \cellcolor{attrcolor!75} 22.00 & \cellcolor{attrcolor!75} 17.67 & \cellcolor{attrcolor!75} 27.03 & \cellcolor{attrcolor!75} 45.00 (\textbf{\textcolor{blue}{+23.00}}) & \cellcolor{attrcolor!75} 12.33 (\textcolor{DarkRed}{\textminus 5.33}) & \cellcolor{attrcolor!75} 15.18 (\textbf{\textcolor{DarkRed}{\textminus 11.85}}) \\

\textbf{} & \textsc{Input×Grad} & \textsc{} & \cellcolor{attrcolor!75} 22.67 & \cellcolor{attrcolor!75} 16.33 & \cellcolor{attrcolor!75} 28.83 & \cellcolor{attrcolor!75} \underline{46.00} (\textbf{\textcolor{blue}{+23.33}}) & \cellcolor{attrcolor!75} \underline{14.00} (\textcolor{DarkRed}{\textminus 2.33}) & \cellcolor{attrcolor!75} \underline{18.75} (\textbf{\textcolor{DarkRed}{\textminus 10.08}}) \\

\grayruletwonine \textbf{} & \textsc{Random} & \textsc{} &  18.00 &  12.67 &  \textbf{\textcolor{DarkRed}{4.50}} &  28.33 (\textbf{\textcolor{blue}{+10.33}}) &  8.67 (\textcolor{DarkRed}{\textminus 4.00}) &  0.89 (\textcolor{DarkRed}{\textminus 3.61}) \\

\textbf{} & \textsc{Human} & \textsc{} &  24.00 &  \textbf{\textcolor{DarkRed}{1.33}} &  40.54 &  51.00 (\textbf{\textcolor{blue}{+27.00}}) &  18.00 (\textbf{\textcolor{blue}{+16.67}}) &  22.32 (\textbf{\textcolor{DarkRed}{\textminus 18.22}}) \\

\textbf{} & \textsc{All} & \textsc{} &  79.00 &  \textbf{\textcolor{DarkRed}{1.33}} &  67.57 &  81.33 (\textcolor{blue}{+2.33}) &  29.67 (\textbf{\textcolor{blue}{+28.33}}) &  73.21 (\textcolor{blue}{+5.65}) \\





\bottomrule
\end{tabular}

\end{center}
\captionsetup{aboveskip=5pt}
\caption{
Faithfulness \textsc{Flip Rate$\uparrow$} percentage. The difference in faitfhulness between the fine-tuned (10 epochs) and pre-trained model is reported in the parentheses. The number of words to mask is enforced (\textsc{Top-Var}), and no method could mask more than the specified number for each sentence (except \textsc{ALL}). The highest faithfulness in each combination of model and dataset is \underline{underlined}. Faithfulness evaluation before fine-tuning is unreliable due to collapsing predictions and biasing toward specific labels which causes \textcolor{DarkRed}{Less than 5\%} flip rate in some cases. Attribution-based methods are more faithful in all of the cases after fine-tuning when the evaluation is more reliable.
}
\label{tab:faithfulness_pt_ft}
\end{table*}

\subsection{Faithfulness to the Model}

While human alignment provides a useful measure of the plausibility of LLM rationales, it is more important to consider the faithfulness of these rationales to the model's actual decision-making process. A word may be crucial for the model's decision even if it does not align with human rationale and vice versa. Therefore, we must ask: Are the self-explanations genuinely influential in the model's decision-making process?

To evaluate faithfulness, we employ a perturbation-based experiment similar to previous work \citep{madsen2024selfexplanations, modarressi-etal-2023-decompx}. In this experiment, we mask the important words identified by the prompting and attribution methods and measure the flip rate of the predicted label during classification. A higher flip rate indicates that the masked words are indeed important to the model, leading it to change its previous decision, and this suggests that the explanation is more faithful to the model's decision-making process. 

\subsubsection{Limitations of Faithfulness Evaluation before Fine-Tuning}

Table~\ref{tab:faithfulness_pt_ft} presents the faithfulness flip rate of the pre-trained and fine-tuned LLMs. 
A noteworthy finding is that in the e-SNLI and FEVER datasets, where classification accuracy was notably low (Figure~\ref{fig:accuracy_epochs_horizontal}), both attribution-based and prompting-based methods result in a very small (Less than 5\%) flip rate in pre-trained models. Even more concerning, masking all the words in the input sentence led to less than a 2\% flip rate for the Mistral and Llama-2 models (Mask \textsc{All}).

This issue arises from pre-trained models being heavily biased toward specific labels in poorly performing datasets (Figure~\ref{fig:labels_distribution}). For instance, in the e-SNLI dataset, Llama2, Llama3, and GPT-3.5 predominantly predict “entailment” in over 80\% of the examples. 
Consequently, even masking the entire input does not change their biased predictions, resulting in extremely low faithfulness flip rates.

\begin{figure*}[t!]
\centering

    \centering
    \includegraphics[width=.32\textwidth]{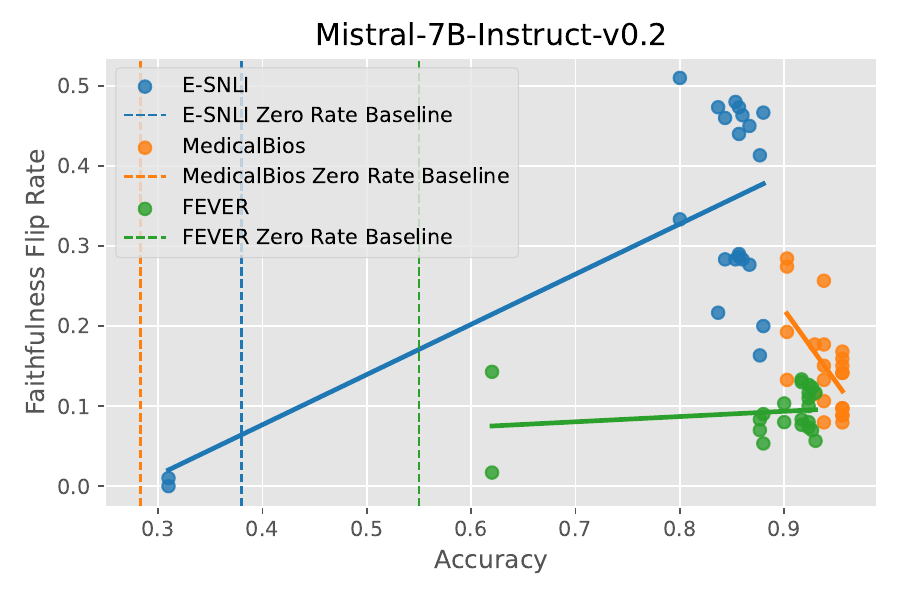}\hfill
    \includegraphics[width=.32\textwidth]{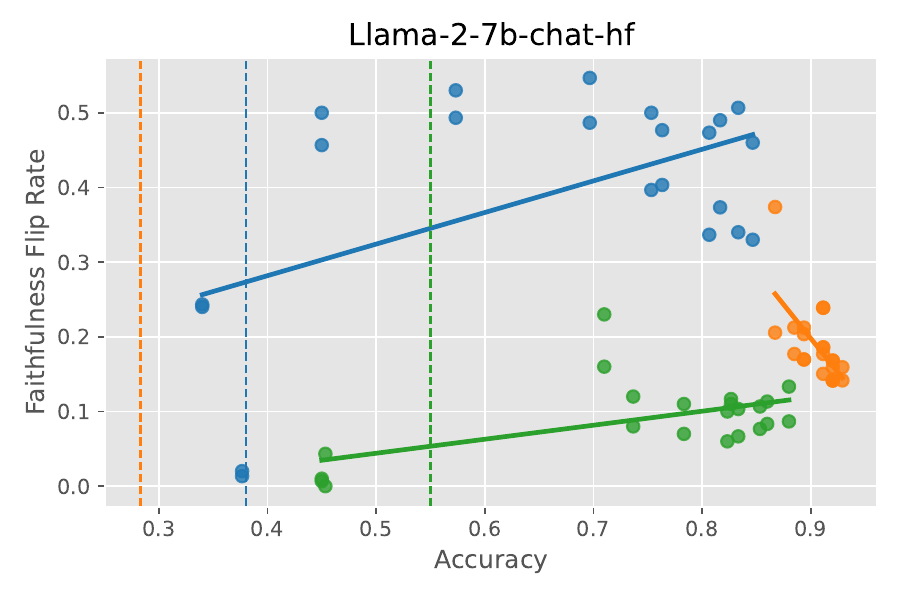}\hfill
    \includegraphics[width=.32\textwidth]{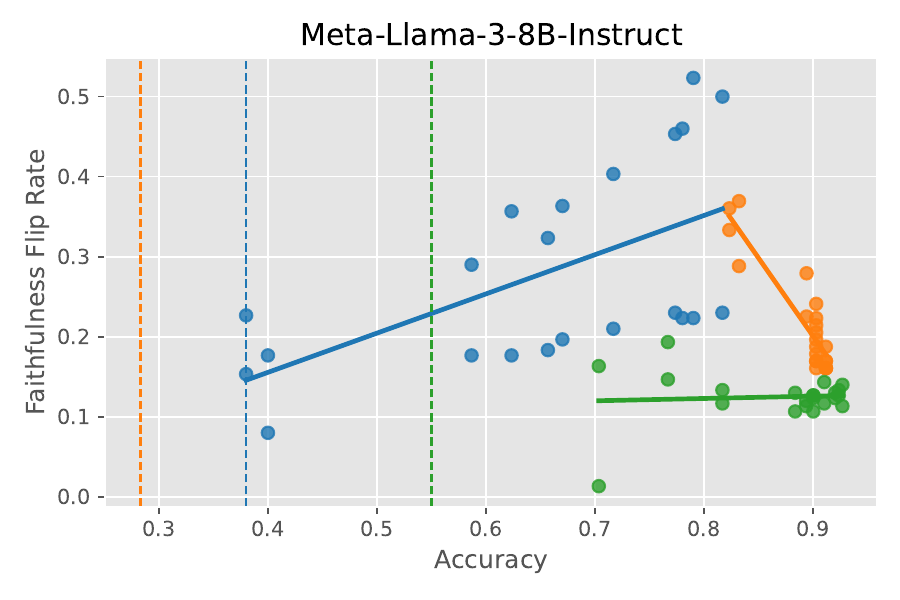}
    \captionsetup{aboveskip=5pt}
    \caption{
    Accuracy and Faithfulness Flip Rate of 10 epochs of fine-tuning.
    }
    \label{fig:acc_faithfulness_corr}
\end{figure*}

Figure~\ref{fig:acc_faithfulness_corr} illustrates the relationship between accuracy and faithfulness flip rate, based on 10 epochs of fine-tuning each model on each dataset. Additionally, Table~\ref{tab:acc_faithfulness_corr} displays the correlation between model faithfulness and classification accuracy over the same 10 epochs, alongside the pre-trained model's accuracy. Both analyses indicate that models and datasets with lower pre-trained performance, relative to the zero-rate baseline, tend to show a stronger correlation between improvements in accuracy and increases in faithfulness. This phenomenon can also be attributed to label bias. When a pre-trained model has very low accuracy, it often indicates significant label bias, meaning that masking any number of words may not impact the biased decision. Fine-tuning the model helps mitigate this label bias (Figure~\ref{fig:labels_distribution}), leading to a higher faithfulness flip rate and, consequently, a correlation with accuracy improvements. 

Although fine-tuning is not a definitive solution, it brings attention to a critical issue in previous faithfulness evaluations of LLMs, which were primarily developed with encoder-based models like BERT in mind. Encoder-based models are typically fine-tuned for specific classification tasks, whereas LLMs are pre-trained for broader tasks such as language modeling and instruction following. As a result, using the same faithfulness evaluations assesses the explanations over language modeling capabilities of LLMs rather than classification tasks, where word attributions can differ significantly. The assumption underlying these evaluations is that if a rationale is truly faithful, masking it should meaningfully change the model's decision.
However, our research shows that for LLMs such as Llama2 and datasets like e-SNLI, the model’s performance can be so poor (e.g., consistently predicting the same class) that even masking the entire input text has little to no impact on its predictions, leading to extremely low faithfulness. This problem is further exacerbated by the differences between BERT's straightforward classifier approach and the prompting used for LLMs, which introduces additional prompt engineering variations and complicates the evaluation process. This performance issue likely led to conclusions in other papers that faithfulness is highly model- and task-dependent. Thus, our study argues that the traditional BERT-based faithfulness evaluation methods are inadequate for current LLMs. Fine-tuning is one approach to addressing this evaluation issue, nonetheless, further research is needed to develop more accurate methods for evaluating the faithfulness of LLM rationales.

\newcolumntype{\pc}{>{\columncolor{promptcolor}}r}
\newcolumntype{\ac}{>{\columncolor{attrcolor}}r}

\begin{table}[t!]
\begin{center}
\small

\tabcolsep=0.11cm

\begin{tabular}{llrr}
\toprule
 &  & PT Acc. & Correlation \\
Dataset & Model &  &  \\
\midrule
\midrule \multirow[c]{3}{*}{\text{E-SNLI}} & \text{Mistral-7B} & 31.00 & 0.68 (0.00) \\
\text{} & \text{Llama-2-7b} & 37.67 & 0.53 (0.01) \\
\text{} & \text{Llama-3-8B} & 38.00 & 0.58 (0.00) \\

\midrule \multirow[c]{3}{*}{\text{FEVER}} & \text{Mistral-7B} & 62.00 & 0.19 (0.41) \\
\text{} & \text{Llama-2-7b} & 45.33 & 0.55 (0.01) \\
\text{} & \text{Llama-3-8B} & 70.33 & 0.07 (0.77) \\

\midrule \multirow[c]{3}{*}{\text{MedicalBios}} & \text{Mistral-7B} & 90.27 & -0.60 (0.00) \\
\text{} & \text{Llama-2-7b} & 86.73 & -0.65 (0.00) \\
\text{} & \text{Llama-3-8B} & 83.19 & -0.91 (0.00) \\

\bottomrule
\end{tabular}

\end{center}
\captionsetup{aboveskip=5pt}
\caption{
The ``Pearson Correlation (p-value)'' between Faithfulness Flip Rate and Classification Accuracy over 10 fine-tuning epochs. Pre-trained accuracy ("PT Acc.") is also reported. Models and datasets with lower pre-trained performance show a stronger correlation between faithfulness and accuracy gains.
}
\vspace{-0.2in}
\label{tab:acc_faithfulness_corr}
\end{table}

\subsubsection{Faithfulness after Fine-Tuning}
Table~\ref{tab:faithfulness_pt_ft} also displays the faithfulness flip rate of the fine-tuned models. 
A comparison of results after fine-tuning (where the near-zero flip rate issue is addressed) reveals that attribution methods outperform prompting methods in faithfulness. 
This difference can stem from attribution methods basing their explanations on the model's internal processes, whereas prompting may provide plausible answers without direct access to this information, potentially diverging from the truth of the model's inner workings. Additionally, prompting is affected by the model's ability to follow instructions, which may result in the generation of an inaccurate number of words or the inclusion of words not present in the input sentence, leading to less faithful results.

Another finding from Table~\ref{tab:faithfulness_pt_ft} and Figure~\ref{fig:acc_faithfulness_corr} is that for MedicalBios, where the model was already performing well, further fine-tuning leads to a decrease in faithfulness. This can occur because additional fine-tuning allows the model to learn more cues from the training set, making it less sensitive to word masking. As a result, masking the same number of words may no longer flip its predictions, since the model can rely more effectively on signals from the unmasked portions. 

Finally, we present the flip rate after masking human rationales in Table~\ref{tab:faithfulness_pt_ft}.
Despite expectations that the model would better recognize the importance of words for its own decisions, these methods consistently underperform human rationales. This result emphasizes that while the current methods demonstrate a degree of faithfulness, there remains room for further refinement and enhancement.

\section{Conclusions}
In this study, we investigated extracting rationales from LLMs, focusing on human alignment and model faithfulness. We experimented with prompting and attribution methods across different LLM architectures and datasets.
Our study shows that attribution methods generally provide rationales that align more closely with human reasoning compared to prompting methods, especially after fine-tuning.
Of greater importance, attribution methods also outperform prompting approaches in faithfulness, as they more accurately reflect the model's internal decision-making process.
Moreover, we showed that fine-tuning reduces label bias and correlates with faithfulness in low-accuracy datasets, explaining the model and task-dependent results of previous works.
Despite these improvements, a gap persists between the models' rationales and human rationales in alignment and faithfulness, underscoring the need for the development of more advanced explanation methods to bridge this gap.

\newpage

\section*{Limitations}
\paragraph{LLM instruction-following abilities.}
In our implementation of prompting strategies, we heavily rely on the LLM's capability to follow instructions accurately. For example, when requesting the top-$k$ words separated by a specific delimiter character, we expect the model to output a list of words in our desired format and quantity with no extra explanations. However, LLMs are still not fully adept at adhering to prompts precisely \cite{sun-etal-2023-evaluating}, which can lead to outputs in various formats different from our expectations. Since our primary focus in this paper is not to evaluate the format-following ability of LLMs, we have taken measures to address discrepancies in the outputs as much as possible.

To mitigate these discrepancies, we adopt tailored parsing approaches to handle unexpected output formats. For instance, if a model separates words in the output with a ``,'' character instead of the instructed character ``|'', we adjust our parsing method accordingly. Fortunately, each model tends to adhere to a relatively consistent output format across the dataset, which enables us to adapt our parsing approach accordingly. Nonetheless, it's worth noting that an LLM with enhanced instruction-following abilities could potentially yield even better parsing results and consequently achieve higher performance levels.

\paragraph{Attribution-based methods}
In selecting the explanation methods based on the inner workings of the models we opted for the ones that were already implemented for LLMs and were relatively efficient to execute given the large size of the models. Nonetheless, we acknowledge that recent vector-based methods have shown promising faithfulness results by decomposing the representations \cite{kobayashi-etal-2020-attention-norm, kobayashi-etal-2021-incorporating-residual, modarressi-etal-2022-globenc, ferrando-etal-2022-measuring, modarressi-etal-2023-decompx} on smaller models such as BERT \cite{devlin-etal-2019-bert} compared with the gradient-based methods. Our study highlights the gap that could be filled by implementing these methods for LLMs.

\paragraph{Prompt Engineering}
Although we reported various versions of prompts for extracting rationales in this paper and conducted preliminary prompt engineering, we acknowledge that better prompts could potentially achieve higher performance. However, this approach diverges from realistic use cases where users may ask questions in various wordings. This limitation is inherent to prompting methods, whereas attribution-based methods are not susceptible to this issue. Therefore, addressing this limitation calls for continued exploration and refinement of both prompting and attribution-based methods in rationale extraction.

\paragraph{Larger Models}
In our experiments, we evaluated open models with less than 8B parameters due to resource limitations. However, we acknowledge that larger models could potentially perform better in following instructions, leading to improved human alignment and model faithfulness in their self-explanations.

\paragraph{Perturbation-based faithfulness evaluation}
In this paper, we conduct faithfulness evaluation of LLM rationales using perturbation-based metrics. Those metrics assume that removing critical features based on rationales would largely affect model performance. However, Whether perturbation-based metrics truly reflect rationale faithfulness is a widely discussed but unsolved question, as they would produce out-of-distribution counterfactuals. For example,~\citet{yin2022sensitivity} show that with different kinds of perturbations such as removal or noise in hidden representations, the faithful sets vary significantly. For consistency, we follow previous work~\citep{deyoung2019eraser, huang2023can}. We leave deeper study into faithfulness measurements of LLM rationales to future work.



\bibliography{anthology, custom}

\appendix
\counterwithin{figure}{section}
\counterwithin{table}{section}
\section{Appendix}
\label{sec:appendix}

\begin{figure*}[t!]
\centering
    \includegraphics[width=0.99\textwidth, trim=0 0 0 0, clip]{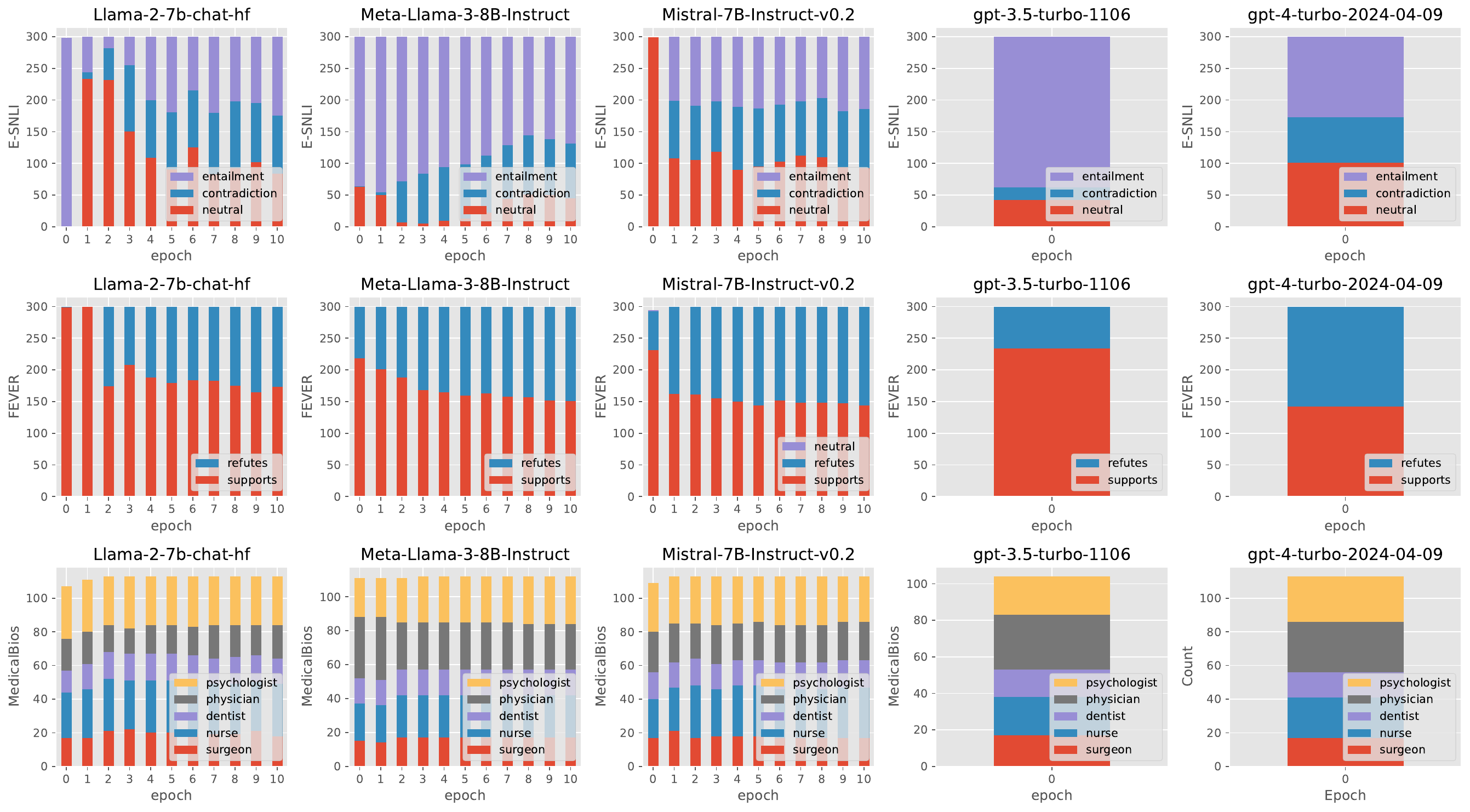}
    \caption{
    The distribution of predicted labels across epochs of fine-tuning. Pre-trained off-the-shelf models tend to be heavily biased toward a label in poorly performing datasets.
    }
    \label{fig:labels_distribution}
\end{figure*}
\begin{figure*}[t!]
\centering
    \includegraphics[width=0.90\textwidth, trim=5 410 0 0, clip]{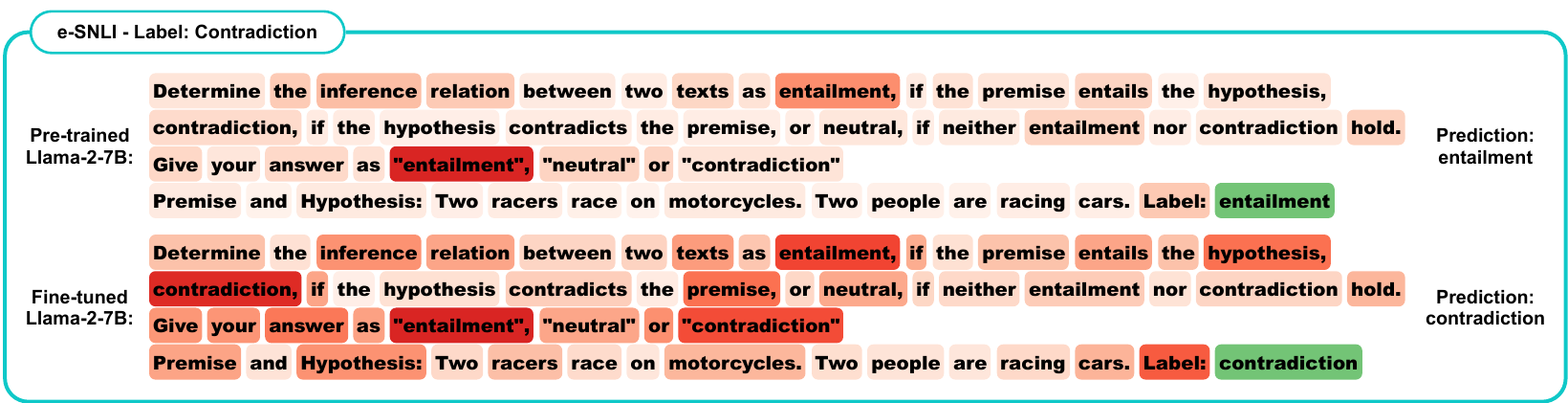}
    \caption{
    Token Importance before and after fine-tuning Llama-2-7B based on the InputXGradient explainability method. The predicted word by the model is shown in green and its attributions to previous words are shown in red. The attributions before fine-tuning are more skewed (Fisher-Pearson coefficient of skewness over all the dataset: 3.37±0.31), and become less skewed after fine-tuning (1.42±0.36).
    }
    \label{fig:token_importance}
\end{figure*}
\begin{figure*}[t!]
\centering
    \includegraphics[width=0.90\textwidth, trim=5 440 0 0, clip]{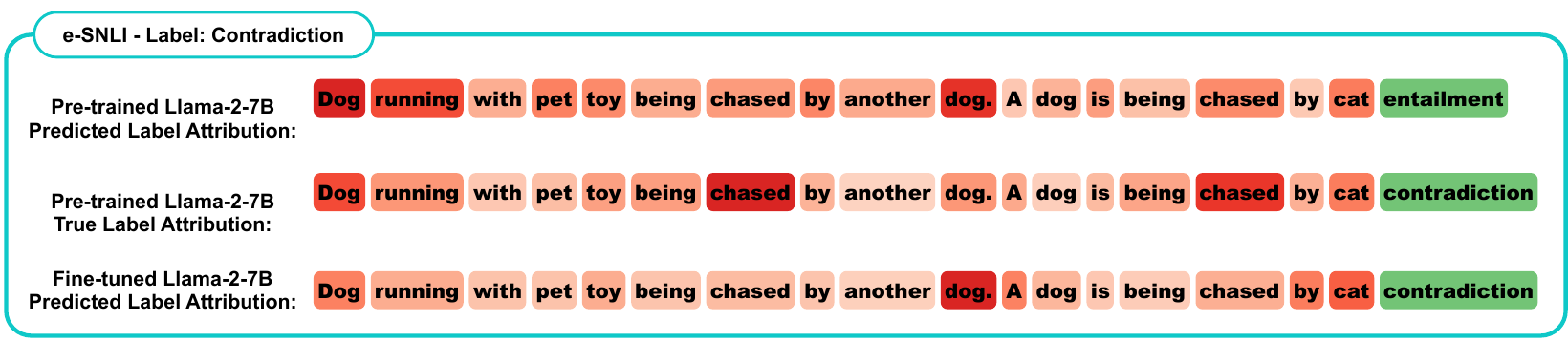}
    \includegraphics[width=0.90\textwidth, trim=5 420 0 0, clip]{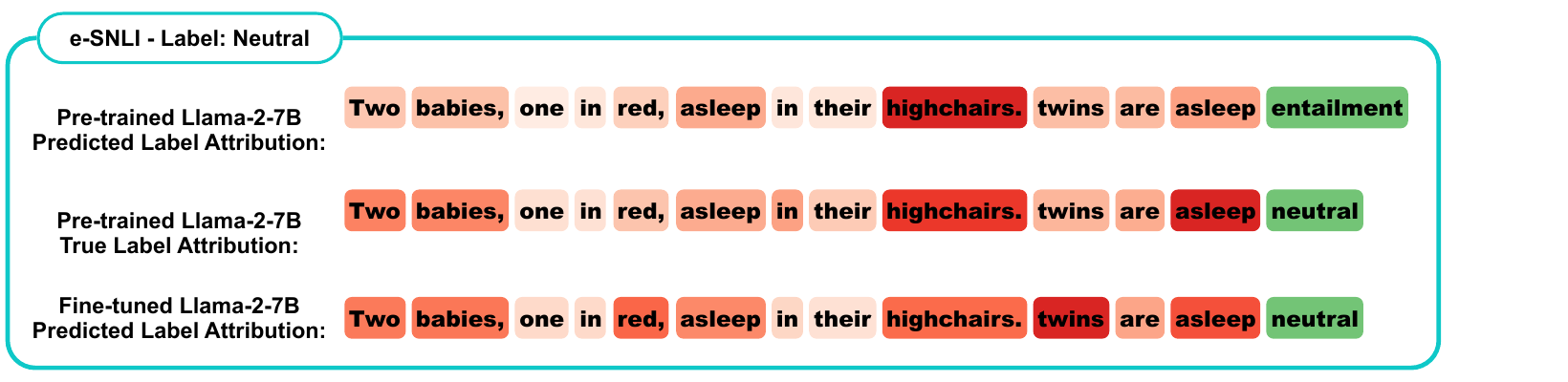}
    \includegraphics[width=0.90\textwidth, trim=5 420 0 0, clip]{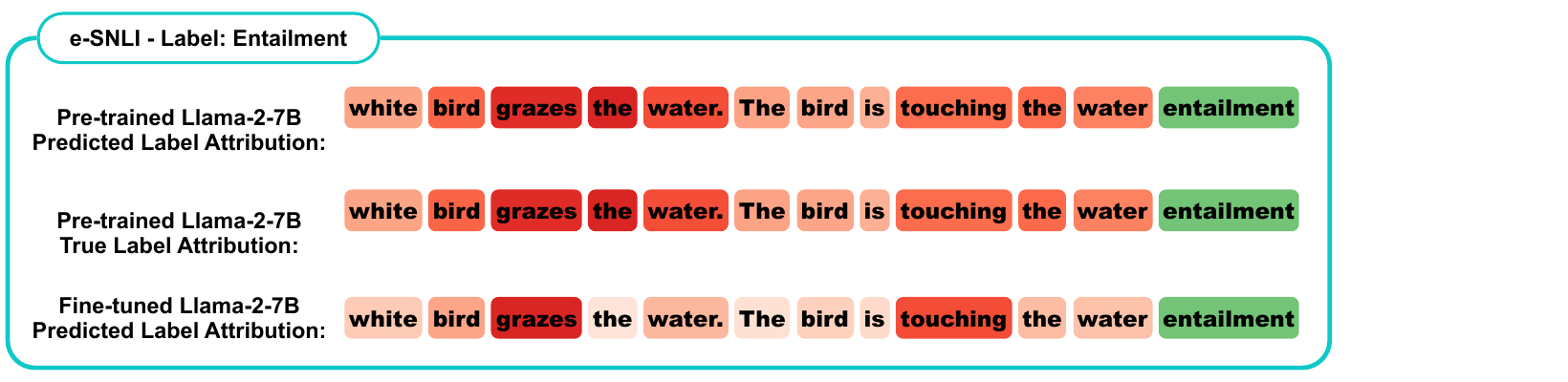}
    \caption{
    Token Importance before and after fine-tuning Llama-2-7B based on the Input×Gradient explainability method. The predicted/true label is shown in green, with its attributions to input words in red. The human rationale for the examples are ["dog", "cat], ["twins"], and ["grazes", "touching"]. The fine-tuned model identified these words solely through training on classification data, without any rationale data.
    }
    \label{fig:token_importance_ft}
\end{figure*}
\begin{table}[h]
\small
\centering
\begin{tabular}{lc}
    \toprule
    \multicolumn{1}{c} {Model} &
    \multicolumn{1}{c} {Access} \\
    \midrule
    \href{https://huggingface.co/meta-llama/Meta-Llama-3-8B-Instruct}{meta-llama/Meta-Llama-3-8B-Instruct} & Open Source \\
    \href{https://huggingface.co/meta-llama/Llama-2-7b-chat-hf}{meta-llama/Llama-2-7b-chat-hf} & Open Source \\
    \href{https://huggingface.co/mistralai/Mistral-7B-Instruct-v0.2}{mistralai/Mistral-7B-Instruct-v0.2} & Open Source \\
    \href{https://platform.openai.com/docs/models/gpt-3-5-turbo}{gpt-3.5-turbo-1106} & Proprietary \\
    \href{https://platform.openai.com/docs/models/gpt-4-turbo-and-gpt-4}{gpt-4-turbo-2024-04-09} & Proprietary \\
    \bottomrule
\end{tabular}
\caption{The details of the models we used in this work.}
\label{tab:models}
\end{table}
\begin{table}[h]
\small
\centering
\begin{tabular}{lc}
    \toprule
    \multicolumn{1}{c} {Hyperparameter} &
    \multicolumn{1}{c} {Value}
    \\
    \midrule
    Total Batch Size & 64 \\
    Learning Rate E-SNLI & 7e-06 \\
    Learning Rate FEVER & 7e-06 \\
    Learning Rate MedicalBios & 3e-06 \\
    Num Epochs & 10 \\
    Learning Rate Scheduler Warmup Steps & 10 \\
    Training Dataset Size & 5000 \\
    LoRA r & 32 \\
    LoRA alpha & 16 \\
    LoRA drop out & 0.05 \\
    \bottomrule
\end{tabular}
\caption{The hyperparameters used for fine-tuning the models using LoRA.}
\label{tab:finetuning_hyperparameters}
\end{table}

\begin{table*}[h]
\centering
\scriptsize
\begin{tabular}{p{2.6cm}p{2.6cm}p{8.0cm}}
\toprule 
\textbf{Method} & \textbf{Selection} & \textbf{Prompt} \\
\midrule 

PROMPT & UNBOUND
&
Premise: ``` \{premise\} ```\newline
Hypothesis: ``` \{hypothesis\} ```\newline
Label: \{label\}\newline

Identify the most important words from the text that are most relevant to understanding the relationship between the premise and the hypothesis. 
Write the words as a pipe-separated (|) list of words with spaces. Do not output any other text, symbols, or explanations.
\\
\midrule

PROMPT & TOP-VAR
&
Premise: ``` \{premise\} ```\newline
Hypothesis: ``` \{hypothesis\} ```\newline
Label: \{label\}\newline

Identify the top \{k\} most important words from the text that are most relevant to understanding the relationship between the premise and the hypothesis. \newline
Write the top \{k\} words as a pipe-separated (|) list of words with spaces. Do not output any other text, symbols, or explanations.
\\
\midrule

ATTRIBUTION-BASED & TOP-VAR \newline (Selected later from tokens with the highest attribution scores)
&
Premise and Hypothesis: ``` \{premise\} \{hypothesis\} ``` \newline

Determine the inference relation between two (short, ordered) texts as entailment, if the premise entails the hypothesis, contradiction, if the hypothesis contradicts the premise, or neutral, if neither entailment nor contradiction hold. 
Respond with exactly one of the following: "entailment", "contradiction", or "neutral." Do not output any other text, symbols, or explanations—just the label.\newline
Label: \{label\}
\\
\midrule

CLASSIFICATION& 
&
Premise and Hypothesis: ``` \{premise\} \{hypothesis\} ``` \newline

Determine the inference relation between two (short, ordered) texts as entailment, if the premise entails the hypothesis, contradiction, if the hypothesis contradicts the premise, or neutral, if neither entailment nor contradiction hold. 
Respond with exactly one of the following: "entailment", "contradiction", or "neutral." Do not output any other text, symbols, or explanations—just the label.\newline
Label: 
\\

\bottomrule
\end{tabular}
\caption{
The prompts utilized for the e-SNLI dataset.
} 
\label{tab:prompts_esnli}
\end{table*}
\begin{table*}[h]
\centering
\scriptsize
\begin{tabular}{p{2.6cm}p{2.6cm}p{8.0cm}}
\toprule 
\textbf{Method} & \textbf{Selection} & \textbf{Prompt} \\
\midrule 

PROMPT & UNBOUND
&
Claim: \{claim\}\newline
Evidence: ``` \{evidence\} ```\newline
Label: \{label\}\newline

Identify the most important words from the given evidence that are most essential for verifying the factual relationship between the claim and the evidence. 
Provide these words as a pipe-separated (|) list. Do not output any other text, symbols, or explanations.
\\
\midrule

PROMPT & TOP-VAR
&
Claim: \{claim\}\newline
Evidence: ``` \{evidence\} ```\newline
Label: \{label\}\newline

Identify the top \{k\} most important words from the given evidence that are most essential for verifying the factual relationship between the claim and the evidence. 
Provide these top \{k\} words as a pipe-separated (|) list. Do not output any other text, symbols, or explanations.
\\
\midrule

ATTRIBUTION-BASED & TOP-VAR \newline (Selected later from tokens with the highest attribution scores)
&
Claim: \{claim\}\newline
Evidence: ``` \{evidence\} ```\newline

Determine the factual relationship between the claim and the supporting evidence as "supports" if the evidence supports the claim or "refutes" if the evidence refutes the claim.
Respond with exactly one of the following: "supports" or "refutes". Do not output any other text, symbols, or explanations—just the label.
Label: \{label\}
\\
\midrule

CLASSIFICATION& 
&
Claim: \{claim\}\newline
Evidence: ``` \{evidence\} ```\newline

Determine the factual relationship between the claim and the supporting evidence as "supports" if the evidence supports the claim or "refutes" if the evidence refutes the claim.
Respond with exactly one of the following: "supports" or "refutes". Do not output any other text, symbols, or explanations—just the label.
Label: 
\\

\bottomrule
\end{tabular}
\caption{
The prompts utilized for the FEVER dataset.
} 
\label{tab:prompts_fever}
\end{table*}
\begin{table*}[h]
\centering
\scriptsize
\begin{tabular}{p{2.6cm}p{2.6cm}p{8.0cm}}
\toprule 
\textbf{Method} & \textbf{Selection} & \textbf{Prompt} \\
\midrule 

PROMPT & UNBOUND
&
Bio: ``` \{bio\} ```\newline

Identify the most important words from the text that indicate why is the person in the following short bio described as a \{label\}.
Write the key words as a pipe-separated (|) list of single words. Do not output any other text, symbols, or explanations.
\\
\midrule

PROMPT & TOP-VAR
&
Bio: ``` \{bio\} ```\newline

Identify the top \{k\} most important words from the text that indicate why is the person in the following short bio described as a \{label\}.
Write the key words as a pipe-separated (|) list of single words. Do not output any other text, symbols, or explanations.
\\
\midrule

ATTRIBUTION-BASED & TOP-VAR \newline (Selected later from tokens with the highest attribution scores)
&
Bio: ``` \{bio\} ```\newline

Determine the occupation of the person in the short bio. 
Respond with exactly one of the following: (psychologist, surgeon, nurse, dentist, physician). Do not output any other text, symbols, or explanations—just the label.
Occupation: \{label\}
\\
\midrule

CLASSIFICATION& 
&
Bio: ``` \{bio\} ```\newline

Determine the occupation of the person in the short bio. 
Respond with exactly one of the following: (psychologist, surgeon, nurse, dentist, physician). Do not output any other text, symbols, or explanations—just the label.
Occupation: 
\\

\bottomrule
\end{tabular}
\caption{
The prompts utilized for the MedicalBios dataset.
} 
\label{tab:prompts_medicalbios}
\end{table*}

\end{document}